# Node Splitting: A Scheme for Generating Upper Bounds in Bayesian Networks

**Arthur Choi, Mark Chavira** and **Adnan Darwiche**
Computer Science Department
University of California, Los Angeles
Los Angeles, CA 90095
{*aychoi,chavira,darwiche*}*@cs.ucla.edu*

## Abstract

We formulate in this paper the mini-bucket algorithm for approximate inference in terms of exact inference on an approximate model produced by splitting nodes in a Bayesian network. The new formulation leads to a number of theoretical and practical implications. First, we show that branch-and-bound search algorithms that use mini-bucket bounds may operate in a drastically reduced search space. Second, we show that the proposed formulation inspires new mini-bucket heuristics and allows us to analyze existing heuristics from a new perspective. Finally, we show that this new formulation allows mini-bucket approximations to benefit from recent advances in exact inference, allowing one to significantly increase the reach of these approximations.

## 1 INTRODUCTION

Probabilistic reasoning tasks in Bayesian networks are typically NP–hard, and approximation algorithms are often sought to address this apparent intractability. One approach to approximate inference is based on *mini-buckets*, a scheme that has been successfully employed by branch-and-bound algorithms for computing MPEs (Most Probable Explanations) (Dechter & Rish, 2003; Marinescu, Kask, & Dechter, 2003). Roughly speaking, mini-buckets is a greedy approach to approximate inference that applies the variable elimination algorithm to a problem, but only as long as computational resources allow it. When time and space constraints keep us from progressing, a mini-buckets approach will heuristically ignore certain problem dependencies, permitting the process of variable elimination to continue (Zhang & Poole, 1996; Dechter, 1996). Mini-buckets will therefore give rise to a family of approximations that, in particular, are guaranteed to produce upper bounds on the value we seek, and further whose quality depends on the heuristic used to ignore dependencies.

In this paper, we make explicit in the most fundamental terms the dependencies that mini-bucket approximations ignore. In particular, we reformulate the mini-bucket approximation using *exact* inference on an approximate *model*, produced by removing dependencies from the original model. We refer to this process of removing dependencies as *node splitting,* and show that any mini-bucket heuristic can be formulated as a node splitting heuristic.

This perspective on mini-buckets has a number of implications, both theoretical and practical. First, it shows how one can significantly reduce the search space of brand-and-bound algorithms that make use of mini-bucket approximations for generating upper bounds. Second, it provides a new basis for designing mini-bucket heuristics, a process which is now reduced to specifying an approximate model that results from node splitting. We will indeed propose a new heuristic and compare it to an existing heuristic, which we reformulate in terms of node splitting. Third, it allows one to embed the mini-bucket approximation in the context of any exact inference algorithm—for example, ones that exploits local structure (Chavira & Darwiche, 2006)—which could speed up the process of generating mini-bucket bounds, without affecting the quality of the approximation. We will illustrate this ability in some of the experiments we present later.

This paper is organized as follows. In Section 2, we review the MPE task, as well as algorithms for finding MPEs. In Section 3, we define node splitting operations for Bayesian networks, and show in Section 4 how mini-bucket elimination is subsumed by splitting nodes. In Section 5, we examine mini-buckets as a node splitting strategy, and introduce a new strategy based on jointrees. In Section 6, we consider branch-and-bound search for finding MPEs, and show how we can exploit node splitting to improve the efficiency of search. In Section 7, we provide empirical support for the claims in Section 6, and conclude in Section 8. Proofs and other results appear in the Appendix.

## 2  MOST PROBABLE EXPLANATION

We will ground our discussions in this paper using the problem of computing MPEs, which we define formally next. Let $N$ be a Bayesian network with variables $\mathbf{X}$, inducing distribution $Pr$. The *most probable explanation* (MPE) for evidence $\mathbf{e}$ is then defined as:

$$MPE(N, \mathbf{e}) \stackrel{def}{=} \arg\max_{\mathbf{x} \sim \mathbf{e}} Pr(\mathbf{x}),$$

where $\mathbf{x} \sim \mathbf{e}$ means that instantiations $\mathbf{x}$ and $\mathbf{e}$ are compatible: they agree on every common variable. Note that the MPE solution may not be unique, in which case $MPE(N, \mathbf{e})$ denotes a set of MPEs. One can also define the *MPE probability*:

$$MPE_p(N, \mathbf{e}) \stackrel{def}{=} \max_{\mathbf{x} \sim \mathbf{e}} Pr(\mathbf{x}).$$

A number of approaches have been proposed to tackle the MPE problem, when a Bayesian network has a high treewidth. These include methods based on local search (Park, 2002; Hutter, Hoos, & Stützle, 2005) and max-product belief propagation (e.g., Pearl, 1988; Weiss, 2000), including generalizations (e.g., Yedidia, Freeman, & Weiss, 2005; Dechter, Kask, & Mateescu, 2002) and related methods (Wainwright, Jaakkola, & Willsky, 2005; Kolmogorov & Wainwright., 2005). Although these approaches have been successful themselves, and can provide high-quality approximations, they are in general non-optimal.

An approach based on systematic search can be used to identify provably optimal MPE solutions, although the efficiency of a search depends heavily on the problem formulation as well as the accompanying heuristics. In particular, it is quite common also to use branch-and-bound search algorithms for computing MPEs and their probability (e.g., Marinescu et al., 2003; Marinescu & Dechter, 2005). The use of these search algorithms, however, requires the computation of an upper bound on the MPE probability to help in pruning the search space. The mini-buckets method is the state of the art for computing such bounds (Dechter & Rish, 2003). In fact, the success of mini-buckets is most apparent in this context of computing MPEs, which is the reason we will use this application to drive our theoretical analysis and empirical results.

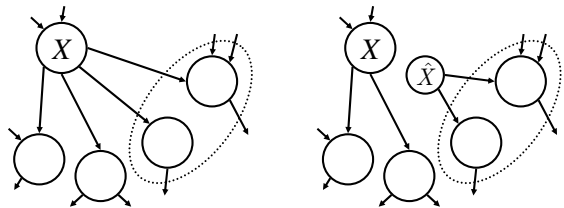

Figure 1: When we split a variable $X$ (left), we create a clone $\hat{X}$ that inherits some of the children (right).

## 3  SPLITTING NODES

We will define in this section a method for approximating Bayesian networks by splitting nodes: An operation that creates a clone $\hat{X}$ of some node $X$, where the clone inherits some of the children of $X$; see Figure 1.

**Definition 1** *Let $X$ be a node in a Bayesian network $N$ with children $\mathbf{Y}$. We say that node $X$ is split according to children $\mathbf{Z} \subseteq \mathbf{Y}$ when it results in a network that is obtained from $N$ as follows:*

- *The edges outgoing from node $X$ to its children $\mathbf{Z}$ are removed.*
- *A new root node $\hat{X}$ with a uniform prior is added to the network with nodes $\mathbf{Z}$ as its children.*

A special case of node splitting is edge deletion, where a node is split according to a single child (i.e., splitting also generalizes edge deletion as defined in Choi & Darwiche, 2006a, 2006b).

**Definition 2** *Let $X \to Y$ be an edge in a Bayesian network $N$. We say that node $X$ is split along an edge $X \to Y$ when the node $X$ is split according to child $Y$.*

The following case of node splitting will be the basis of a splitting strategy that yields a special class of mini-bucket approximations with implications in search.

**Definition 3** *Let $X$ be a node in a Bayesian network $N$. We say that node $X$ is fully split when $X$ is split along every outgoing edge $X \to Y$.*

Thus, when we fully split a node $X$, we create one clone for each of its outgoing edges. Figure 2 illustrates an example of a network where two nodes have been split. Node $C$ has been split according to children $\{D, E\}$, and Node $A$ has been split along the edge $A \to D$.

A network $N'$ which results from splitting nodes in network $N$ has some interesting properties. To explicate these properties, however, we need to introduce a function which, given an instantiation $\mathbf{x}$ of variables in network $N$, gives us an instantiation of clones in $N'$ that agrees with the values given to variables in $\mathbf{x}$.

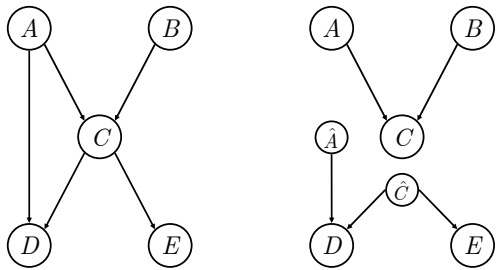

Figure 2: A Bayesian network $N$ (left) and an approximation $N'$ (right) found by splitting $C$ according to $\{D, E\}$, and splitting $A$ according to $D$.

**Definition 4** *Let $N$ be a Bayesian network, and let $N'$ be the result of splitting nodes in $N$. If $\mathbf{x}$ is an instantiation of variables in $N$, then let $\overrightarrow{\mathbf{x}}$ be the compatible instantiation of the corresponding clones in $N'$.*

For example, in the split network in Figure 2, an instantiation $\mathbf{x} = \{A=a_1, B=b_1, C=c_2, D=d_3, E=e_1\}$ is compatible with instantiation $\overrightarrow{\mathbf{x}} = \{\hat{A}=a_1, \hat{C}=c_2\}$. Moreover, $\mathbf{x}$ is not compatible with $\{\hat{A}=a_1, \hat{C}=c_1\}$.

To see the effect that splitting a node can have on a network, consider a simple two-node network $A \to B$ with binary variables, where $\theta_{a_1} = .2$, $\theta_{b_1|a_1} = .1$, and $\theta_{b_1|a_2} = .7$. After splitting $A$ according to $B$, we have:

| $\mathbf{x}$ | $Pr(\mathbf{x})$ | $\mathbf{x}'$ | $Pr'(\mathbf{x}')$ | $\mathbf{x}'$ | $\beta Pr'(\mathbf{x}')$ |
|---|---|---|---|---|---|
| $a_1 b_1$ | 0.02 | $a_1 \hat{a}_1 b_1$ | 0.01 | $a_1 \hat{a}_1 b_1$ | 0.02 |
| $a_1 b_2$ | 0.18 | $a_1 \hat{a}_1 b_2$ | 0.09 | $a_1 \hat{a}_1 b_2$ | 0.18 |
| $a_2 b_1$ | 0.56 | $a_1 \hat{a}_2 b_1$ | 0.07 | $a_1 \hat{a}_2 b_1$ | 0.14 |
| $a_2 b_2$ | 0.24 | $a_1 \hat{a}_2 b_2$ | 0.03 | $a_1 \hat{a}_2 b_2$ | 0.06 |
| | | $a_2 \hat{a}_1 b_1$ | 0.04 | $a_2 \hat{a}_1 b_1$ | 0.08 |
| | | $a_2 \hat{a}_1 b_2$ | 0.36 | $a_2 \hat{a}_1 b_2$ | 0.72 |
| | | $a_2 \hat{a}_2 b_1$ | 0.28 | $a_2 \hat{a}_2 b_1$ | 0.56 |
| | | $a_2 \hat{a}_2 b_2$ | 0.12 | $a_2 \hat{a}_2 b_2$ | 0.24 |

where $\beta = |A_1| = 2$. We see that whenever $A_1$ and its clone $\hat{A}_1$ are set to the same value, we can recover the original probabilities $Pr(\mathbf{x})$ after splitting, by using $\beta Pr'(\mathbf{x}')$. This includes the value of the MPE in $N$, which may no longer be the largest value of $\beta Pr'(\mathbf{x}')$.

This intuition yields the key property of split networks.

**Theorem 1** *Let $N$ be a Bayesian network, and let $N'$ be the result of splitting nodes in $N$. We then have*

$$MPE_p(N, \mathbf{e}) \leq \beta MPE_p(N', \mathbf{e}, \overrightarrow{\mathbf{e}}).$$

*Here, $\beta = \prod_{C \in \mathbf{C}} |C|$, where $\mathbf{C}$ is the set of clones in network $N'$.*

That is, the MPE probability with respect to a split network provides an upper bound on the MPE probability with respect to the original network. We note that the probability of evidence is also upper bounded in the split network; see Theorem 3 in the Appendix.

---

**Algorithm 1** VE($N, \mathbf{e}$): returns $MPE_p(N, \mathbf{e})$.

1: $i \leftarrow 0$
2: $\mathcal{S} \leftarrow \{f^{\mathbf{e}} \mid f^{\mathbf{e}} \text{ is a CPT (incorporating } \mathbf{e}\text{) of } N\}$
3: **while** $\mathcal{S}$ contains variables **do**
4: $\quad i \leftarrow i + 1$
5: $\quad X \leftarrow$ a variable appearing in $\mathcal{S}$
6: $\quad \mathcal{S}_i \leftarrow$ **all** factors in $\mathcal{S}$ that contain $X$
7: $\quad f_i \leftarrow \max_X \prod_{f \in \mathcal{S}_i} f$
8: $\quad \mathcal{S} \leftarrow \mathcal{S} - \mathcal{S}_i \cup \{f_i\}$
9: **return** product of factors in $\mathcal{S}$

---

**Algorithm 2** MBE($N, \mathbf{e}$): returns an upper bound on $MPE_p(N, \mathbf{e})$.

{ Identical to Algorithm 1, except for Line 6: }
6: $\quad \mathcal{S}_i \leftarrow$ **some** factors in $\mathcal{S}$ that contain $X$

---

The following corollary shows that splitting degrades the quality of approximations monotonically.

**Corollary 1** *Let network $N_2$ be obtained by splitting nodes in network $N_1$, which is obtained by splitting nodes in network $N_0$. We then have*

$$\begin{aligned} MPE_p(N_0, \mathbf{e}) &\leq \beta_1 MPE_p(N_1, \mathbf{e}, \overrightarrow{\mathbf{e}_1}) \\ &\leq \beta_2 MPE_p(N_2, \mathbf{e}, \overrightarrow{\mathbf{e}_2}), \end{aligned}$$

*where $\beta_1, \beta_2$ and $\overrightarrow{\mathbf{e}_1}, \overrightarrow{\mathbf{e}_2}$ are as defined by Theorem 1.*

## 4 MINI-BUCKET ELIMINATION

We discuss in this section the relationship between the approximations returned by split networks and those computed by the mini-buckets algorithms (Dechter & Rish, 2003). In particular, we show that every mini-buckets heuristic corresponds precisely to a node splitting strategy, where exact inference on the resulting split network yields the approximations computed by mini-buckets. Our discussion here will be restricted to computing MPEs, yet the correspondence extends to probability of evidence as well.

We start first by a review of the mini-buckets method, which is a relaxed version of the variable elimination method given in Algorithm 1 (Zhang & Poole, 1996; Dechter, 1996). According to this algorithm, variable elimination starts with a set of factors corresponding to the CPTs of a given Bayesian network. It then iterates over the variables appearing in factors, eliminating them one at a time. In particular, to eliminate a variable $X$, the method multiplies all factors that contain $X$ and then max-out $X$ from the result. The bottleneck of this algorithm is the step where the factors containing $X$ are multiplied, as the resulting factor may be too big for the computational resources available. The mini-bucket method deals with this difficulty by making a simple change to the variable elimination algorithm (also known as the bucket elimination algorithm).[1] This change concerns Line 6 in which all factors containing variable $X$ are selected. In mini-buckets, given in Algorithm 2 (Dechter & Rish, 2003), one chooses only a subset of these factors in order to control the size of their product. Which particular set of factors is chosen depends on the specific heuristic used. Yet, regardless of the heuristic used, the answer obtained by the mini-buckets method is guaranteed to be an upper bound on the correct answer.[2] One should note here that the simple change from **all** to **some** on Line 6 implies the following. The number of iterations performed by Algorithm 1 is exactly the number of network variables, since each iteration will eliminate a network variable. However, Algorithm 2 may only partially eliminate a variable in a given iteration, and may take multiple iterations to eliminate it completely.

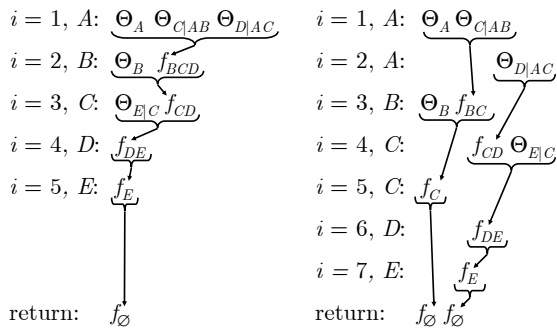

Figure 3: An execution trace of VE on $N$ (left) and MBE on $N$ (right). The network is defined in Figure 2.

To help us visualize the computations performed by Algorithms 1 and 2, consider their execution trace.

**Definition 5** *Given an instance of* VE *or* MBE *run on a given network $N$, we define its execution* trace *$\mathcal{T}$ as a labeled DAG which adds, for each iteration $i$,*

- *a node $i$, labeled by the factor set $\mathcal{S}_i$, and*

- *directed edges $j \to i$, for all factors $f_j \in \mathcal{S}_i$, each labeled by the corresponding factor $f_j$.*

---

[1]More precisely, bucket elimination is a particular implementation of variable elimination in which one uses a list of buckets to manage the set of factors during the elimination process. Although the use of such buckets is important for the complexity of the algorithm, we ignore them here as the use of buckets is orthogonal to our discussion.

[2]This is also true for versions of the algorithm that compute the probability of evidence.

Figure 3 depicts traces of both algorithms on the network in Figure 2 (left). Variable elimination, whose trace is shown on the left, eliminates variables from $A$ to $E$, and performs five iterations corresponding to the network variables. Mini-buckets, however, performs seven iterations in this case, as it takes two iterations to eliminate variable $A$ and two iterations to eliminate variable $C$. Note that an execution trace becomes a rooted tree after reversing the direction of all edges.

Given an execution trace $\mathcal{T}$, we can visually identify all of the network CPTs used to construct any factor in Algorithms 1 and 2. For mini-buckets, we also want to identify a subtrace of $\mathcal{T}$, but one that covers only those network CPTs that are relevant to a particular *attempt* at eliminating variable $X$ at iteration $i$. A CPT is *not* relevant to iteration $i$ if $X$ is eliminated from it in a later iteration, or if $X$ has already been eliminated from it in some previous iteration.

Given a trace $\mathcal{T}$, we thus define the *subtrace* $\mathcal{T}_i$ relevant to an iteration $i$ as the nodes and edges of $\mathcal{T}$ that are reachable from node $i$ (including itself), but only by walking up edges $j \to i$, and only those edges labeled with factors $f_j$ mentioning variable $X$. For example, in Figure 3 (right), the subtrace $\mathcal{T}_i$ for iteration $i = 7$ is the chain $4 \to 6 \to 7$. In the same trace, the subtrace $\mathcal{T}_i$ for iteration $i = 5$ is the chain $1 \to 3 \to 5$.

Given a subtrace $\mathcal{T}_i$, we can identify only those CPTs that are relevant to a partial elimination of $X$, but further, the set of variables those CPTs belong to.

**Definition 6** *Let $i$ be an iteration of* MBE *where we eliminate variable $X$, and let $\mathcal{T}_i$ be the subtrace of $\mathcal{T}$ that is relevant to iteration $i$. The* <u>basis</u> **B** *of an iteration $i$ is a set of variables where $Y \in \mathbf{B}$ iff:*

- $\Theta_{Y|\mathbf{U}} \in \mathcal{S}_j$ *for some node $j$ of $\mathcal{T}_i$, and*

- $X \in \{Y\} \cup \mathbf{U}$,

*where $\Theta_{Y|\mathbf{U}}$ are CPTs in $N$.*

For example, in Figure 3 (right), the basis of iteration $i = 4$ is $\{D, E\}$, since $C$ is eliminated from the CPTs of $D$ and $E$ at iteration 4.

Given this notion, we can show how to construct a network with split nodes, that corresponds to a particular execution of the mini-bucket method. In particular, exact variable elimination in $N'$ will be able to mimic mini-bucket elimination in $N$, with the same computational complexity. This is given in Algorithm 3 which returns both a network $N'$ and an ordering $\pi'$ of the variables in $N'$ (this includes the variables in original network $N$ and their clones in $N'$). Figure 4 shows a trace corresponding to a split network, and the associated variable order.

**Algorithm 3** SPLIT-MBE($N, \mathbf{e}$): returns a split network $N'$ and variable ordering $\pi'$, corresponding to a run of MBE($N, \mathbf{e}$).

1: $N' \leftarrow N$
2: **for** each iteration $i$ of MBE($N, \mathbf{e}$) **do**
3:    $X \leftarrow$ as chosen on Line 5 of MBE
4:    $\mathcal{S}_i \leftarrow$ as chosen on Line 6 of MBE
5:    $\mathbf{B} \leftarrow$ basis of iteration $i$
6:    **if** $X \in \mathbf{B}$ **then**
7:      $\pi'(i) \leftarrow X$
8:    **else**
9:      split node $X$ in $N'$ according to children $\mathbf{B}$
10:      $\pi'(i) \leftarrow$ clone $\hat{X}$ of $X$ resulting from split
11: **return** network $N'$ and ordering $\pi'$

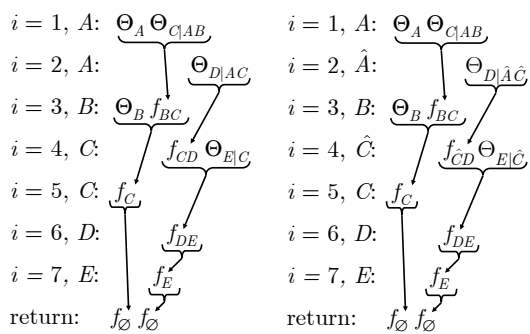

Figure 4: An execution trace of MBE on $N$ (left) and VE on $N'$ (right). For simplicity, we ignore the priors of clone variables in $N'$. Networks are defined in Figure 2.

We now have our basic correspondence between mini-buckets and node splitting.

**Theorem 2** *Let $N$ be a Bayesian network, $\mathbf{e}$ be some evidence, and let $N'$ and $\pi'$ be the results of* SPLIT-MBE($N, \mathbf{e}$). *We then have:*

$$\text{MBE}(N, \mathbf{e}) = \beta MPE_p(N', \mathbf{e}, \overrightarrow{\mathbf{e}}),$$

*where $\beta = \prod_{C \in \mathbf{C}} |C|$ and $\mathbf{C}$ are the clone variables in $N'$. Moreover, variable elimination on network $N'$ using the variable order $\pi'$ has the same time and space complexity of the corresponding run* MBE($N, \mathbf{e}$).

Note that the ordering $\pi'$ returned by Algorithm 3 may not be the most efficient ordering to use when running exact variable elimination in a split network: there may be another variable order where VE($N', \mathbf{e}, \overrightarrow{\mathbf{e}}$) produces smaller intermediate factors than MBE($N, \mathbf{e}$). Indeed, we need not restrict ourselves to variable elimination when performing inference on the split network, as any exact algorithm suffices for this purpose. This property can have significant practical implications, a point we highlight in Section 7 where we exploit recent advances in exact inference algorithms.

## 5 NODE-SPLITTING STRATEGIES

Given the correspondences in the previous section, every mini-bucket heuristic can now be interpreted as a node splitting strategy. Consider for example the mini-bucket heuristic given in (Dechter & Rish, 2003), which is a greedy strategy for bounding the size of the factors created by MBE. This heuristic works as follows, given a bound on the size of the largest factor:

- A particular variable order is chosen and followed by the heuristic.

- When processing variable $X$, the heuristic will pick a maximal set of factors $\mathcal{S}_i$ whose product will be a factor of size within the given bound.

- The above process is repeated in consecutive iterations and for the same variable $X$ until variable $X$ is eliminated from all factors.

- Once $X$ is completely eliminated, the heuristic picks up the next variable in the order and the process continues.

This heuristic tries then to minimize the number of instances where a proper subset of factors is selected in Line 6 of Algorithm 2, and can be interpreted as a heuristic to minimize the number of clones introduced into an approximation $N'$. In particular, the heuristic does not try to minimize the number of split variables.

We now introduce a new node splitting strategy based on *fully* splitting nodes, where a variable is split along every outgoing edge. The strategy is also a greedy algorithm, which attempts to fully split the variable that contributes most to the difficulty of running a *jointree* algorithm in the approximate network $N'$. This process is repeated until the network is sufficiently simplified. In particular, the method starts by building a jointree of the original network. It then picks a variable whose removal from the jointree will introduce the largest reduction in the sizes of the cluster and separator tables. Once a variable is chosen, it is fully split. One can obtain a jointree for the split network by simply modifying the existing jointree, which can then be used to choose the next variable to split on.[3] In our empirical evaluation, we go further and construct a new jointree for the simpler network, and choose the next variable to split from it. This process is repeated until the largest jointree cluster is within our bound.

We now have two strategies for splitting nodes in a network. The first is based on the classical mini-bucket heuristic that tries to minimize the number of clones,

---

[3]In particular, one can simply adjust the separators and clusters without changing the structure of the jointree.

**Algorithm 4** SPLIT-BNB: $\mathbf{z}$ and $q^\star$ are global variables.

1: $q \leftarrow \beta MPE_p(N', \mathbf{z}, \overrightarrow{\mathbf{z}})$
2: **if** $q > q^\star$ **then**
3:   **if** $\mathbf{z}$ is a complete instantiation **then**
4:     $q^\star \leftarrow q$
5:   **else**
6:     pick some $X \notin \mathbf{Z}$
7:     **for** each value $x$ of variable $X$ **do**
8:       $\mathbf{z} \leftarrow \mathbf{z} \cup \{X = x\}$
9:       SPLIT-BNB()
10:      $\mathbf{z} \leftarrow \mathbf{z} - \{X = x\}$

and the second one is based on reducing the size of jointree tables and tries to minimize the number of split variables. Recall that Corollary 1 tells us that the quality of the MPE bound given by a split network degrades monotonically with further splits. As we shall see in Section 6, and empirically in Section 7, it may sometimes be more important to minimize the number of split variables, rather than the number of clones, in the context of branch-and-bound search.

## 6 SEARCHING FOR MPE'S

When computing the MPE is too difficult for traditional inference algorithms, we can employ systematic search methods to identify provably optimal solutions.

Suppose now that we are given network $N$ and evidence $\mathbf{e}$, and that we want to compute $MPE_p(N, \mathbf{e})$ using depth-first branch-and-bound search. We want then to select some network $N'$ using a node-splitting heuristic from the previous section to allow for exact inference in $N'$ (say, by the jointree algorithm). Theorem 1 gives us the upper bound

$$MPE_p(N, \mathbf{e}) \leq \beta MPE_p(N', \mathbf{e}, \overrightarrow{\mathbf{e}}).$$

Moreover, one can easily show that if $\mathbf{z}$ is a complete variable instantiation $\mathbf{x}$ of $N$, we then have

$$MPE_p(N, \mathbf{x}) = \beta MPE_p(N', \mathbf{x}, \overrightarrow{\mathbf{x}});$$

see Lemma 1. These two properties form the basis of our proposed search algorithm, SPLIT-BNB, which is summarized in Algorithm 4.

Throughout the search, we keep track of two global variables. First, $\mathbf{z}$ is a partial assignment of variables in the original network that may be extended to produce an MPE solution in $MPE(N, \mathbf{e})$. Second, $q^\star$ is a lower bound on the MPE probability that is the largest probability of a complete instantiation so far encountered. The search is initiated after setting $\mathbf{z}$ to $\mathbf{e}$ and $q^\star$ to 0.0: we use evidence $\mathbf{e}$ as the base instantiation, and 0.0 as a trivial lower bound. Upon completion of the search, we have the optimal MPE probability $q^\star = MPE_p(N, \mathbf{e})$.

At each search node, we compute a bound on the best completion of $\mathbf{z}$ by performing exact inference in the approximate network $N'$. If the resulting upper bound $q$ is greater than the current lower bound $q^\star$, then we must continue the search, since it is possible that $\mathbf{z}$ can provide us with a better solution than what we have already found. In this case, if $\mathbf{z}$ is already a complete instantiation, it is easy to show that $q$ is equal to $Pr(\mathbf{z})$ (by Lemma 1, in the Appendix) and that we have found a new best candidate solution $q^\star$. If $\mathbf{z}$ is not a complete instantiation, we select some variable $X$ that has not been instantiated. For each value $x$ of $X$, we add the assignment $\{X = x\}$ to $\mathbf{z}$ and call SPLIT-BNB recursively with the new value of $\mathbf{z}$ and our candidate solution $q^\star$. Upon returning from the recursive call, we retract the assignment $\{X = x\}$, and continue to the next value of $X$.

### 6.1 REDUCING THE SEARCH SPACE

Consider now the following critical observation.

**Proposition 1** *Let $N$ be a Bayesian network, and let $N'$ be the result of splitting nodes in $N$. If $\mathbf{Z}$ contains all variables that were split in $N$ to produce $N'$, then*

$$MPE_p(N, \mathbf{z}) = \beta MPE_p(N', \mathbf{z}, \overrightarrow{\mathbf{z}}),$$

*where $\beta = \prod_{C \in \mathbf{C}} |C|$ and $\mathbf{C}$ are all the clones in $N'$.*

According to this proposition, once we have instantiated in $\mathbf{z}$ all variables that were cloned, the resulting approximation is exact. This tells us that during our search, we need not instantiate every one of our network variables $\mathbf{X}$. We need only instantiate variables in a smaller set of variables $\mathbf{Z} \subseteq \mathbf{X}$ containing precisely the variables that were split in $N$ to produce $N'$. Once the bound on the MPE probability becomes exact, we know that we will not find a better solution by instantiating further variables, so we can stop and backtrack. This observation allows us to work in a reduced search space: rather than searching in a space whose size is exponential in the number of network variables $\mathbf{X}$, we search in a space whose size is exponential only in the number of split variables!

Moreover, if our variable splitting strategy seeks to minimize the number of split variables, rather than the number of clones introduced, we can potentially realize dramatic reductions in the size of the resulting search space. As we shall see in the following section, this can have a drastic effect on the efficiency of search.

## 7 EMPIRICAL OBSERVATIONS

We present empirical results in this section to highlight the trade-offs in the efficiency of search based on the quality of the bound resulting from different node splitting strategies, and the size of the resulting search space. We further illustrate how our framework allows for significant practical gains with relatively little effort, by employing state-of-the-art algorithms for exact inference in the approximate, node-split network. Thus, our goal here is, not to evaluate a completely specified system for MPE search, but to illustrate the benefits that our node-splitting perspective can bring to existing systems.

We begin with experiments on networks for decoding error-correcting codes (see, e.g., Frey & MacKay, 1997; Rish, Kask, & Dechter, 1998). We first consider simpler networks, that correspond to codes containing 16 information bits and 24 redundant bits. Each of our plot points is an average of 42 randomly generated networks: 6 networks for each of 7 levels of noise.[4] Here, an MPE solution would recover the most likely word encoded prior to transmission. Our method for exact inference in the approximate model is based on compiling Bayesian networks (Chavira & Darwiche, 2007), an approach that has already been demonstrated to be effective in branch-and-bound search for MAP explanations (Huang, Chavira, & Darwiche, 2006).

In our experiments, we compared the splitting strategy based on a jointree (JT) with the strategy based on a greedy mini-bucket elimination (MB), both described in Section 5. In particular, we asserted limits on the maximum cluster size for JT, and equivalently, the size of the largest factor for MB. We then compared the two strategies across a range of cluster and factor size limits from 0 to 12, where 0 corresponds to a fully disconnected network and 12 corresponds to exact inference (no splits). In all of our experiments, to emphasize the difference between splitting strategies, we make neutral decisions in the choice of a search seed (we use a trivial seed, 0.0), variable ordering (random) and value ordering (as defined by the model).

First, consider Figure 5, which compares the effectiveness of node splitting strategies in minimizing the number of variables split and the number of clones. Recall that the heuristic based on jointrees (JT) seeks to minimize the number of split variables, while the greedy mini-bucket (MB) strategy would seek to minimize the number of clones. We see that in Figure 5, on the left, our jointree (JT) method can split nearly half of the variables that the mini-bucket (MB) strategy splits. On the other hand, we see that on the right, the mini-bucket (MB) strategy is introducing fewer clones. Note that on both extremes (no splits and all split), MB and JT are identical.

---
[4]In particular, each network is associated with its own piece of evidence corresponding to a codeword received via transmission through a (simulated) noisy Gaussian channel, with standard deviations ranging from $\sigma = 0.2$ to $\sigma = 0.8$ in steps of 0.1.

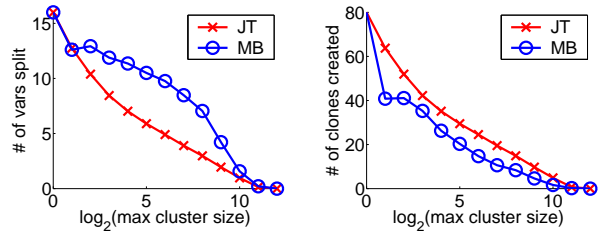

Figure 5: Comparing splitting heuristics.

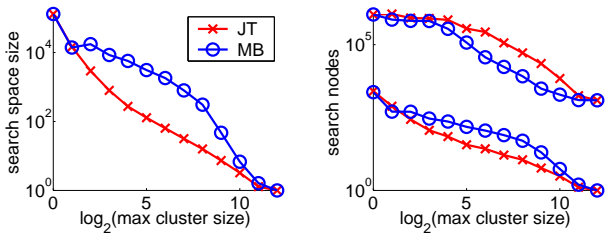

Figure 6: Evaluating the efficiency of search. On the right, the top pair searches in the full space, and the bottom pair searches in the reduced space.

To see the impact that reducing the number of split variables has on the efficiency of search, consider Figure 6. On the left, we see that JT can get an order of magnitude savings over MB in the size of the reduced search space, which is exponential only in the number of split variables (see again Figure 5). Consider now, on the right, the number of nodes visited while performing SPLIT-BNB search. The top pair plots the efficiency of search using the full search space (JT-F and MB-F), while the bottom pair plots the efficiency of using the reduced search space (JT-R and MB-R). We see that both JT-R and MB-R experience several orders of magnitude improvement when using the reduced-search space versus the full search space.

When we compare JT-F and MB-F (top pair), we see that MB-F is in fact more efficient in terms of the number of nodes visited. In this setting, where both methods are searching in the same space, we see that the number of clones introduced appears to be the dominant factor in the efficiency of search. This is expected, as we expect that the upper bounds on the MPE probability should be tighter when fewer clones are introduced. When we now compare JT-R and MB-R (bottom pair), we see that the situation has reversed, and that JT-R is now outperforming MB-R. Here, each method is performing search in their own reduced search spaces. A strategy based on reducing the number of split variables reduces the size of the search space, and this reduction now dominates the quality of the bound.

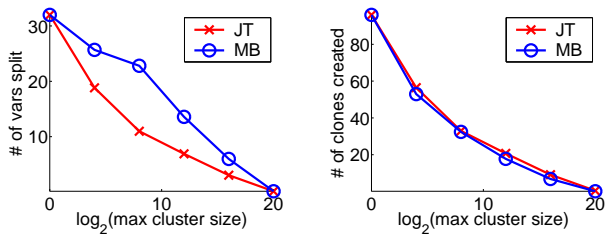

Figure 7: Comparing splitting heuristics.

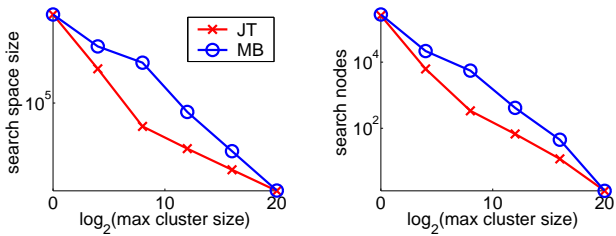

Figure 8: Evaluating the efficiency of search.

Table 1: Compilation versus Variable Elimination

| Network | Search Nodes | AC Time (s) | VE Time (s) | Imp. |
|---|---|---|---|---|
| 90-20-1 | 14985 | 18 | 2417 | 135 |
| 90-20-2 | 137783 | 111 | 15953 | 144 |
| 90-20-3 | 3065 | 4 | 1271 | 334 |
| 90-20-4 | 4545 | 3 | 988 | 355 |
| 90-20-5 | 29343 | 38 | 6579 | 173 |
| 90-20-6 | 5065 | 3 | 630 | 227 |
| 90-20-7 | 2987 | 2 | 1155 | 485 |
| 90-20-8 | 6213 | 6 | 812 | 146 |
| 90-20-9 | 5121 | 5 | 2367 | 480 |
| 90-20-10 | 8419 | 10 | 2343 | 235 |

Figures 7 and 8 depict similar results but for larger coding networks, in which we have a rate $\frac{1}{2}$ code with 32 information bits and 32 redundant bits. Note that only the reduced space was used for search here.

Our approach based on node splitting has another major advantage, which we have only briefly mentioned thus far. By formulating mini-buckets as exact inference in an approximate network, the evaluation of the mini-bucket approximation need not rely on any specific exact inference algorithm. We mention here that the arithmetic circuit (AC) approach we have been using to compute the bound indeed has a key advantage over mainstream algorithms, in that it is able to effectively exploit certain types of local structure (Chavira & Darwiche, 2006). To highlight the extent to which using a different algorithm can be significant, we constructed another set of experiments. In each, we used a different grid network, first introduced in (Sang, Beame, & Kautz, 2005), and constructed a single MPE query. Each grid network has treewidth in the low thirties, just out of reach for traditional algorithms for exact inference. We ran our search twice, each time using a different algorithm to compute the mini-bucket bound: the first using AC and the second using standard variable elimination (that does not exploit local structure). Table 1 shows the results for each network, including the number of search nodes visited and, for each algorithm, the total search time. For each network, we performed two identical searches for each algorithm: the only difference being in *how the bound was computed*. Consequently, the dramatic differences we observe reflect the ability of the AC approach to exploit local structure, showing how advances in exact inference can be easily utilized to extend the reach of mini-bucket approximations.

## 8    CONCLUSION

We presented in this paper a new perspective on mini-bucket approximations, formulating it in terms of exact inference in an approximate network, but one found by splitting nodes. This perspective has led to a number of theoretical and practical insights. For one, it becomes apparent that a branch-and-bound search using a mini-bucket bound may operate in a drastically reduced search space. This suggests a heuristic for identifying a mini-bucket approximation that is explicitly based on minimizing this search space, rather than the quality of the resulting bound. Empirically, we observe that a reduced search space can have more impact than a better bound, in terms of the efficiency of branch-and-bound search. Moreover, as our approach is independent of the algorithm used for exact inference in the resulting approximate network, we can effortlessly employ state-of-the-art algorithms for exact inference, including those that can exploit compilation and local structure.

## A    PROOFS

**Lemma 1** *Let $N$ be a Bayesian network, and let $N'$ be the result of splitting nodes in $N$. We then have*

$$Pr(\mathbf{x}) = \beta Pr'(\mathbf{x}, \overrightarrow{\mathbf{x}}).$$

Here $\beta = \prod_{C \in \mathbf{C}} |C|$, where $\mathbf{C}$ is the set of clones in network $N'$.

**Proof of Lemma 1** Note first that $\overrightarrow{\mathbf{x}}$ is an instantiation of only root variables, and that all clones have uniform priors, i.e., $\theta_c = |C|^{-1}$. We then have that

$$Pr'(\overrightarrow{\mathbf{x}}) = \prod_{c \sim \overrightarrow{\mathbf{x}}} \theta_c = \prod_{C \in \mathbf{C}} |C|^{-1} = \beta^{-1}.$$

Since instantiation $\mathbf{x}$ is compatible with $\overrightarrow{\mathbf{x}}$, where a variable and its clones are set to the same value, we find in $Pr'(\mathbf{x} \mid \overrightarrow{\mathbf{x}})$ that clone variables act as selectors for the CPT values composing $Pr(\mathbf{x})$. Thus

$$Pr'(\mathbf{x}, \overrightarrow{\mathbf{x}}) = Pr'(\mathbf{x} \mid \overrightarrow{\mathbf{x}})Pr'(\overrightarrow{\mathbf{x}}) = Pr(\mathbf{x})\beta^{-1}$$

and we have $Pr(\mathbf{x}) = \beta Pr'(\mathbf{x}, \overrightarrow{\mathbf{x}})$, as desired. $\square$

**Proof of Theorem 1** Suppose for contradiction that there exists an instantiation $\mathbf{z} \in MPE(N, \mathbf{e})$ such that $Pr(\mathbf{z}) > \beta MPE_p(N', \mathbf{e}, \overrightarrow{\mathbf{e}})$. By Lemma 1, the instantiation $\overrightarrow{\mathbf{z}}$ gives us

$$Pr(\mathbf{z}) = \beta Pr'(\mathbf{z}, \overrightarrow{\mathbf{z}}) > \beta MPE_p(N', \mathbf{e}, \overrightarrow{\mathbf{e}}),$$

contradicting the optimality of $MPE_p(N', \mathbf{e}, \overrightarrow{\mathbf{e}})$. $\square$

Proposition 1 is in fact a generalization of Lemma 1 from a complete instantiation $\mathbf{x}$ to a partial instantiation $\mathbf{z}$ where $Z$ contains all nodes that have been split in $N'$. Note that splitting a node $X$ when the value of $X$ has already been fixed corresponds to a common preprocessing rule for Bayesian networks given evidence. In particular, when a given piece of evidence $\mathbf{z}$ fixes the value of variable $Z$, any edge $Z \to Y$ can be pruned and a selector node $\hat{Z}$ can be made a parent of $Y$. Node $\hat{Z}$ is then set to the value that instantiation $\mathbf{z}$ assigns to $Z$. This pruning process yields a simpler network which corresponds exactly to the original network for any query of the form $\{\alpha, \mathbf{z}\}$.

**Proof of Proposition 1** From the correspondence to pruning edges outgoing instantiated variables, we know that queries of the form $\{\alpha, \mathbf{z}\}$, including complete instantiations $\{\mathbf{x}, \mathbf{z}\}$, are equivalent in $N$ conditioned on $\mathbf{z}$ and $N'$ conditioned on $\{\mathbf{z}, \overrightarrow{\mathbf{z}}\}$. Thus the MPEs of each network must also be the same. $\square$

**Proof of Theorem 2** Given the *trace* of an instance of MBE$(N, \mathbf{e})$, algorithm SPLIT-MBE$(N, \mathbf{e})$ returns a network $N'$ and an ordering $\pi'$ of variables in $N'$. We show, by induction, that each iteration of VE on $N'$ mimics each iteration of MBE on $N$. We can then conclude that the product of factors returned by both must be the same, and further, that they are of the same time and space complexity. In particular, we show how VE$(N, \mathbf{e}, \overrightarrow{\mathbf{e}})$ mimics MBE$(N, \mathbf{e})$ first on Line 2, and then Lines 5, 6 and 7, in Algorithms 1 and 2. For simplicity, we ignore the constant factor $\beta$ that the clone CPTs contribute to the MPE value of $N'$.

On Line 2 (iteration $i = 0$), by construction, the CPTs in $N$ are the same as the CPTs in $N'$, after relabeling. For iterations $i > 0$, assume for induction that the factors available to both VE and MBE are the same.

On Line 5, if MBE picked variable $X$ on Line 5, then algorithm VE picks variable $X' = \pi'(i)$, which is either $X$ or a clone $\hat{X}$, by construction (Lines 7 and 10 of Algorithm 3).

On Line 6, each factor in the set $\mathcal{S}_i$ is either 1) a CPT mentioning $X$, or 2) a factor that is composed of a CPT mentioning $X$. The variables that these CPTs belong to are the variable set $\mathbf{B}$, the basis of iteration $i$. Algorithm 3 decides to split (or not split), so that each variable in $\mathbf{B}$ will have a CPT in $N'$ that mentions $X' = \pi'(i)$. We know by induction, that all factors $f$ selected by MBE are available for selection by VE in $N'$. Since Algorithm 3 ensures that each of these factors $f$ now mention $X'$, and since VE picks **all** factors mentioning $X'$, we know VE picks the same factors MBE picked.

On Line 7, consider any variable $Z$ mentioned in $\mathcal{S}_i$. Let $j \geq i$ be the iteration where $Z$ is eliminated in MBE. The relevant CPTs mentioning $Z$ at iteration $i$ are among the relevant CPTs of the basis at iteration $j$. Thus, Algorithm 3 ensures that they all mention the same instance of $Z$ in $N'$. Thus, the resulting product of factors $f_i$ must be the same after relabeling. $\square$

A node-split network also upper bounds $Pr(\mathbf{e})$. The following theorem corresponds to a mini-bucket bound on the probability of evidence (Dechter & Rish, 2003).

**Theorem 3** *Let $N$ be a Bayesian network, and let $N'$ be the result of splitting nodes in $N$. We then have*

$$Pr(\mathbf{e}) \leq \beta Pr'(\mathbf{e}, \overrightarrow{\mathbf{e}}).$$

Here, $\beta = \prod_{C \in \mathbf{C}} |C|$, where $\mathbf{C}$ is the set of clones in network $N'$.

**Proof of Theorem 3** By Lemma 1, we know that $Pr(\mathbf{x}) = \beta Pr'(\mathbf{x}, \overrightarrow{\mathbf{x}})$. Therefore:

$$Pr(\mathbf{e}) = \sum_{\mathbf{x} \sim \mathbf{e}} Pr(\mathbf{x}) = \beta \sum_{\mathbf{x} \sim \mathbf{e}} Pr'(\mathbf{x}, \overrightarrow{\mathbf{x}})$$

$$\leq \beta \sum_{\mathbf{x}' \sim \mathbf{e}, \overrightarrow{\mathbf{e}}} Pr'(\mathbf{x}') = \beta Pr'(\mathbf{e}, \overrightarrow{\mathbf{e}})$$

where $\mathbf{x}'$ is an instantiation of variables in $N'$, but where the values of the original network variables are not necessarily compatible with the values of the clone variables (as they are in $\mathbf{x}$ and $\overrightarrow{\mathbf{x}}$). $\square$

## B  LOOP-CUTSET CONDITIONING

The loop-cutset conditioning algorithm and SPLIT-BNB search are closely related when our splitting strategy performs only full splits (see Definition 3). This correspondence reveals the difficulty of answering the following decision problem:

**D-FS:** Given $k$ and $\omega$, does there exist a set $\mathbf{Z}$ of size $\leq k$ such that fully splitting nodes $\mathbf{Z}$ in network $N$ results in an approximate network $N'$ with treewidth $\leq \omega$?

We now state the following negative result.

**Theorem 4** *Decision problem* **D-FS** *is* NP–*complete*.

Hardness can be shown by reduction from the loop-cutset problem, which is NP–complete (Suermondt & Cooper, 1990). In particular, when we fully split enough variables $\mathbf{Z}$ to render $N'$ a polytree, then $\mathbf{Z}$ also constitutes a loop-cutset of $N$.

If $N'$ is rendered a polytree, and we ignore the bound during SPLIT-BNB search and further employ the reduced search space over split variables $\mathbf{Z}$, then SPLIT-BNB reduces to loop-cutset conditioning. More generally, when we split enough variables $\mathbf{Z}$ so that network $N'$ has treewidth $\omega$, SPLIT-BNB reduces to $\omega$–cutset conditioning (Bidyuk & Dechter 2004).

Assuming that for exact inference in $N'$, we use an algorithm that is exponential in the treewidth $\omega$ of $N'$, this correspondence tells us that the worst-case time and space complexity of SPLIT-BNB search is precisely that of $\omega$–cutset conditioning. In particular, say that $n$ is the number of variables in $N$, value $m$ is the number of variables cloned in $N'$, and value $\omega$ is the treewidth of network $N'$. The worst-case time complexity of SPLIT-BNB search is thus

$$O(n \exp\{\omega\} \cdot \exp\{m\}) = O(n \exp\{\omega + m\}),$$

since we spend $O(n \exp\{\omega\})$ time at each of at most $\exp\{m\}$ search nodes. Note that the space complexity of SPLIT-BNB search is only $O(n \exp\{\omega\} + m)$.